# An Effective Evolutionary Clustering Algorithm: Hepatitis C case study

M. H. Marghny
Computer Science Department, Faculty of Computer and Information, Assiut University, Egypt.

Rasha M. Abd El-Aziz
Computer Science Department, Faculty of Science, Assiut University, Egypt.

Ahmed I. Taloba
Computer Science Department, Faculty of Computer and Information, Assiut University, Egypt.

## ABSTRACT
Clustering analysis plays an important role in scientific research and commercial application. K-means algorithm is a widely used partition method in clustering. However, it is known that the K-means algorithm may get stuck at suboptimal solutions, depending on the choice of the initial cluster centers. In this article, we propose a technique to handle large scale data, which can select initial clustering center purposefully using Genetic algorithms (GAs), reduce the sensitivity to isolated point, avoid dissevering big cluster, and overcome deflexion of data in some degree that caused by the disproportion in data partitioning owing to adoption of multi-sampling.

We applied our method to some public datasets these show the advantages of the proposed approach for example Hepatitis C dataset that has been taken from the machine learning warehouse of University of California. Our aim is to evaluate hepatitis dataset. In order to evaluate this dataset we did some preprocessing operation, the reason to preprocessing is to summarize the data in the best and suitable way for our algorithm. Missing values of the instances are adjusted using local mean method.

## General Terms
Data Mining.

## 1. INTRODUCTION
Data mining is supported by a host of models capture the character of data in several different ways. Clustering is one of these models, where clustering is to find groups that are different from each other's, and whose members are very similar to each other. Clustering is a process in which a group of unlabeled patterns are partitioned into a number of sets so that similar patterns are assigned to the same cluster, and dissimilar patterns are assigned to different clusters. There are two goals for clustering algorithms: determining good clusters and doing so efficiently.

A popular category of clustering algorithms dealing with k-center clustering problems is the so-called K-means clustering. The conventional K-means algorithm classifies data by minimizing the MSE objective function. However, the clustering procedure assumes that data must be presented with a known number of clusters, which in turn allows for a simple implementation of K-means clustering.

However, the classical K-means suffers from several flaws. First, the algorithm is very sensitive on the choice of the initial cluster centers. Second, k-means can't deal with massive data.

Nowadays, people who have been affected from hepatitis virus are found all over the world. Liver's swelling and redness without indicating any particular reason is referred to as Hepatitis [1].It is currently estimated that approximately 170-300 million people worldwide (roughly 3 % of the present global population) are infected with HCV and the global burden of disease attributable to HCV-related chronic liver diseases is substantial [2].

In this paper, we present an improved genetic k-means algorithm (IGK) to handle large scale data, which can select initial clustering center purposefully using genetic algorithms (GAs). GAs based clustering technique provides an optimal clustering with respect to the clustering metric being considered.

A Genetic algorithms (GAs) is a procedure used to find approximate solutions to search problems through the application of the principles of evolutionary biology. Genetic algorithms use biologically inspired techniques, such as genetic inheritance, natural selection, mutation, and sexual reproduction (recombination, or crossover) [3].

An excellent survey of GAs along with the programming structure used can be found in [4]. Genetic algorithms have been applied to many classification and performance tuning applications in the domain of knowledge discovery in databases (KDD).

## 2. CLUSTERING
Cluster analysis, an important technology in data mining, is an effective method of analyzing and discovering useful information from numerous data [5-11]. Cluster algorithm groups the data into classes or clusters so that objects within a cluster have high similarity in comparison to one another, but are very dissimilar to objects in other clusters [12]. Dissimilarities are assessed based on the attribute values describing the objects. Often, distance measures are used. As a branch of statistics and an example of unsupervised learning, clustering provides us an exact and subtle analysis tool from the mathematic view.

The K-means algorithm [12-14] is by far the most popular clustering tool used in scientific and industrial applications [3]. It proceeds as follows. First, it randomly selects k objects, each of which initially represents a cluster mean or center. For each of the remaining objects, an object is assigned to the cluster to which it is the most similar, based on the distance between the object and the cluster mean. It then computes the new mean for each cluster. This process iterates until the criterion function converges. Typically, the squared-error criterion is used, defined as:

$$J_c = \sum_{j=1}^{c} \sum_{k=1}^{n_j} \left\| x_k^j - m_j \right\| \quad (1)$$

Where $J_c$, is the sum of square-error for all objects in the database, c number of clusters, $n_j$ number of objects in each cluster, $x_k$ is the point in space representing a given object, and m is the mean of cluster $c_j$.

Adopting the squared-error criterion, K-means works well when the clusters are compact clouds that are rather well separated from one another and has difficulty detecting the natural clusters, when clusters have widely different sizes, densities, or non-spherical shapes [11].For attempting to minimize the square-error criterion, it will divide the objects in one cluster into two or more clusters. In addition to that, when applying this square-error criterion to evaluate the clustering results, the optimal cluster corresponds to the extremum. Since the objective function has many local minimal values [15], if the results of initialization are exactly near the local minimal point, the algorithm will terminate at a local optimum. So, random selecting initial cluster center is easy to get in the local optimum not the entire optimal.

For overcoming that square-error criterion is hard to distinguish the big difference among the clusters, one technique has been developed which is based on representative point-based technique [16]. Besides, there are various approaches to solving the problem that the performance of algorithm heavily depends on the initial starting conditions: the simplest one is repetition with different random selections [17]; some algorithms also employ simulation anneal technique to avoid getting into local optimal [18]; GA-clustering algorithm [5] and Improved K-means algorithm [11] have been proposed to solve the clustering problem; some algorithms used GAs to solve the clustering problem [19-30]. In literature [31], Bradley and Fayyad present an iterative refinement approach for sampling dataset many times and clustering twice to get the optimal initial values of cluster center. The idea is that multiple sub-samples are drawn from the dataset clustered independently, and then these solutions are clustered again respectively, the refined initial center is then chosen as the solution having minimal distortion over all solutions.

Aiming at the dependency to initial conditions and the limitation of K-means algorithm that applies the square-error criterion to measure the quality of clustering, this paper presents a new improved Genetic K-means algorithm that is based on effective techniques of multi-sampling and once-clustering to search the optimal initial values of cluster centers. Our experimental results demonstrate that the new algorithm can obtain better stability and excel the original K-means in clustering results.

## 3. ORIGINAL K-MEANS CLUSTERING

In this section, we briefly describe the original K-means algorithm.K-means algorithm [15] is the most popularclustering algorithmbut it is suffers from some drawbacks as shown above. The steps ofK-means algorithm as follows:

**Algorithm**: Original K-means(S, k), S= {$x_1$, $x_2$... $x_n$}.

**Input**: The number of clusters k and a dataset containing n objects $x_i$.

**Output**: A set of k clusters $C_j$ that minimize the squared-error criterion.

```
Begin
1. m=1;
2. initialize k prototypes; //arbitrarily chooses k objects as the
initial centers.
3. Repeat
for i=1 to n do
begin
for j=1 to k do
```
Compute $D(X_i, Z_j) = |X_i, Z_j|$ ;//$Z_j$is the center of cluster j.
if $D(X_i, Z_j) = \min\{D(X_i, Z_j)\}$ then
$X_i \in C_j$ ;
end;// (re)assign each object to the cluster based on the mean
if m=1 then
$$J_c(m) = \sum_{j=1}^{k} \sum_{X_i \in C_j} |X_i - Z_j|^2$$
m=m+1;
for j=1 to k do
$$Z_j = \frac{1}{n_j} \sum_{i=1}^{n_j} x_i^{(j)}$$ ; //(re)calculate the mean value of the objectsfor each cluster
$$J_c(m) = \sum_{j=1}^{k} \sum_{X_i \in C_j} |X_i - Z_j|^2$$ ; //compute the error function
4. Until $J_c(m) - J_c(m-1) < \zeta$
End

## 4. GA-CLUSTERING ALGORITHM

**Algorithm**: GA-clustering (S, k), S = {x1, x2,…xn}.

**Input**: The number of clusters K and a dataset containing n objects Xi.
**Output**: A set of k clusters Cj that minimize the squared-error

```
Begin

1. T=0
2. Initialize population P(T)
3. Compute fitness P(T)
4. T=T+1
5. If termination criterion achieved go to 10
6. Select P(T) from P(T-1)
7. Crossover P(T)
8. Mutate P(T)
9. Go to 3
10. Output best and stop

End
```

criterion.

## 5. IMPROVED K-MEANS ALGORITHM

**Algorithm**: Improved K-means (S, k), S = {$x_1, x_2,…,x_n$}.

**Input**: The number of clusters K'(K'>K) and a dataset containing n objects $X_i$.

**Output**: A set of k clusters $C_j$ that minimize the squared-error criterion.

> Begin
> 1. Multiple sub-samples {$S_1, S_2, ...,S_j$};
> 2. For m = 1 to j do
>     K-means($S_m$, K'); //executing K-means, produce K' clusters and j groups.
> 3. Compute $J_c(m) = \sum_{j=1}^{k} \sum_{X_i \in C_j} |X_i - Z_j|^2$ ;
> 4. Choose min{$J_c$} as the refined initial points $Z_j$ , j ∈ [1, K'];
> 5. K-means(S, K'); //executing K-means again with chosen initial, producing K'mediods.
> 6. Repeat
>     Combining two near clusters into one cluster, and recalculate the new center generated by two centers merged.
> 7. Until the number of clusters reduces into k //Merging (K' +K)
> End

## 6. PROPOSED ALGORITHM

Original K-means algorithm choose k points as initial clustering centers, different points may obtain different solutions. In order to diminish the sensitivity of initial point choice, we employ a mediod [12], which is the most centrally located object in a cluster, to obtain better initial centers.

The demand of stochastic sampling is naturally bias the sample to nearly represent the original dataset, that is to say, samples drawn from dataset can't cause distortion and can reflect original data's distribution. The example of original data set is depicted on the left of Fig. 1, while sampling results are shown on the right of Fig. 1.

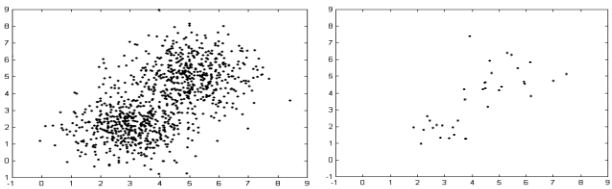

Fig. 1 Dataset and sub-sample.

Comparing two solutions generated by clustering sample drawn from the original dataset and itself using K-means respectively, the location of clustering centroids of these two are almost similar. So, the sample-based method is applicable to refine initial conditions [26]. In order to lessen the influence of sample on choosing initial starting points, following procedures are employed. First, drawing multiple sub-samples (say J) from original dataset (the size of each sub-sample is not more than the capability of the memory, and the sum for the size of J sub-samples is as close as possible to the size of original dataset). Second, use Genetic K-means for each sub-sample and producing a group of mediods respectively. Finally, comparing J solutions and choosing one group having minimal value of square-error function as the refined initial points.

To avoid dividing one big cluster into two or more ones for adopting square-error criterion, we assume the number of clustering is K' (K' >K), K' depends on the balance of clustering quality and time). In general, bigger K' can expand searching area of solution space, and reduce the situation that there are not any initial values near some extremum. Subsequently, re-clustering the dataset through Genetic K-means with the chosen initial conditions would produce K' mediods, then merging K' clusters (which are nearest clusters) until the number of clusters reduced to k.

The mean steps of proposed algorithm can be summarized as follows:

**Algorithm**: Improved Genetic K-means (S, k), S = {$x_1, x_2,…x_n$}.

**Input**: The number of clusters K'(K'> K) and a dataset containing n objects $X_i$.

**Output**: A set of k clusters $C_j$ that minimize the squared-error criterion

> Begin
> 1. Multiple sub-samples {$S_1, S_2, ...,S_j$};
> 2. For m = 1 to j do
>     Genetic K-means($S_m$, K'); //executing Genetic K-means, produce K' clusters and j groups.
> 3. Compute $J_c(m) = \sum_{j=1}^{k} \sum_{X_i \in C_j} |X_i - Z_j|^2$ ;
> 4. Choose min{$J_c$} as the refined initial points $Z_j$ , j ∈ [1, K'];
> 5. Genetic K-means(S, K'); //executing Genetic K-means again with chosen initial, producing K'mediods.
> 6. Repeat
>     Combining two near clusters into one cluster, and recalculate the new center generated by two centers merged.
> 7. Until the number of clusters reduces into k //Merging (K' +K)
> End

Improved algorithm works on very small samples compared with whole dataset and needs significantly less iteration.

## 7. EXPERIMENTAL RESULTS

We run experiments on three real-life data sets. Vowel data, this data consists of 871 Indian Telugu vowel sounds [32]. These were uttered in a consonant-vowel-consonant context by three male speakers in the age group of 30-35 years. The value of K is therefore chosen to be 6 for this data. Iris data, this data represents different categories of irises having four feature values. The four feature values represent the sepal length, sepal width, petal length and the petal width in centimeters [33]. It has three classes (with some overlap between classes 2 and 3) with 50 samples per class. The value of K is therefore chosen to be 3 for this data. Crude oil data, this overlapping data [34] has 56 data points, 5 features and 3 classes. Hence the value of K is chosen to be 3 for this data set.

We also run experiments on three synthetic datasets denoted as A1, A2 and A3 [35], which are shown in Fig. 2.

| Initial | K-means | GA clustering | Improved k-means | IGK |
|---|---|---|---|---|
| 1 | 359.761 | 278.965 | 279.484 | 278.27 |
| 2 | 279.484 | 278.965 | 279.27 | 278.27 |
| 3 | 359.761 | 278.965 | 278.965 | 278.27 |
| 4 | 279.27 | 278.965 | 279.484 | 278.27 |
| 5 | 279.484 | 278.965 | 278.965 | 278.27 |
| Average | 311.552 | 278.965 | 279.2336 | 278.27 |

The Genetic parameters that have been used in our experimental: the population size = 15, selection is roulette introduced in section 2.2.4, crossover is single point crossover demonstrated in section 2.2.5, the probability of crossover = 0.8, mutation is an occasional random alteration of a character in GA clustering algorithm and the probability of mutation = 0.001.

The results of implementation of the K-means algorithm, GA-clustering algorithm, improved K-means algorithm and improved Genetic K-means algorithm are shown, respectively, in Table 1 for vowel dataset, Table 2 for iris dataset , Table 3 for Crude oil dataset , Table 4 for A1 dataset , Table 5 for A2 dataset and Table 6 for A3 dataset. All the algorithms were run for 10 iterations. For the purpose of demonstration, five different initial configurations of the K-means algorithm and improved K-means algorithm and five different initial populations of the GA-clustering algorithm and improved Genetic K-means algorithm are shown in the tables.

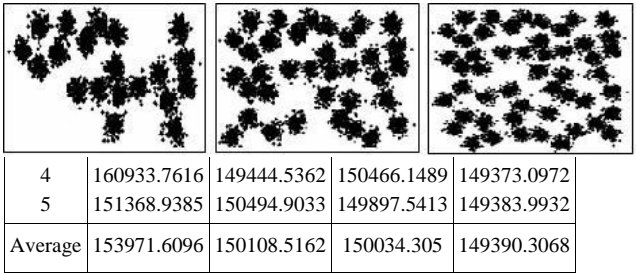

| | | | | |
|---|---|---|---|---|
| 4 | 160933.7616 | 149444.5362 | 150466.1489 | 149373.0972 |
| 5 | 151368.9385 | 150494.9033 | 149897.5413 | 149383.9932 |
| Average | 153971.6096 | 150108.5162 | 150034.305 | 149390.3068 |

Fig. 2 Datasets, A1 (left), A2 (center) and A3 (right).

Table 1: $J_c$ for Vowel after 100 iterations when k = 6.

Table 2: $J_c$ for Iris after 100 iterations when k = 3.

| Initial | K-means | GA clustering | Improved k-means | IGK |
|---|---|---|---|---|
| 1 | 124.022 | 97.1011 | 97.101 | 97.101 |
| 2 | 122.946 | 97.101 | 97.101 | 97.101 |
| 3 | 123.581 | 97.101 | 97.101 | 97.101 |
| 4 | 97.346 | 97.101 | 97.101 | 97.101 |
| 5 | 97.346 | 97.101 | 97.101 | 97.001 |
| Average | 113.0482 | 97.101 | 97.101 | 97.101 |

Table 3: $J_c$ for Crude Oil after 100 iterations when k = 3.

Table 1, 2 and 3 show the calculated error of the K-means algorithm, GA-clustering algorithm, improved K-means algorithm and improved Genetic K-means algorithm, it is obviously that the Squared-error obtained by improved Genetic K-means algorithm for all data set (vowel 149376.3, iris 97.101, crude oil 278.270) is better than the calculated Squared-error by K-means (vowel 155105.5, iris 113.0482, crude oil 311.552), GA-clustering algorithm (vowel 149460.5, iris 97.101, crude oil 278.965) and improved K-means algorithm (vowel 149830, iris 97.101, crude oil 279.2336).

Table 4: $J_c$ for A1 after 10 iterations when k = 20.

| Initial | K-means | GA clustering | Improved k-means | IGK |
|---|---|---|---|---|
| 1 | 7319267.603 | 6290237.938 | 6134273.372 | 5376844.286 |
| 2 | 6939286.418 | 5902763.334 | 7446758.964 | 5376826.391 |
| 3 | 6688417.311 | 6429371.415 | 7471731.766 | 5376844.286 |
| 4 | 6201366.244 | 5955839.693 | 5376844.286 | 5376826.391 |
| 5 | 6732038.312 | 5769628.764 | 6175011.237 | 5376844.286 |
| Average | 6776075.178 | 6069568.229 | 6520923.925 | 5376837.128 |

Table 5: $J_c$ for A2 after 10 iterations when k = 35.

| Initial | K-means | GA clustering | Improved k-means | IGK |
|---|---|---|---|---|
| 1 | 11527418.85 | 9935582.45 | 11979602.2 | 9167097.531 |
| 2 | 11793057.34 | 10462638.41 | 11013299.08 | 9563340.229 |
| 3 | 12219094.33 | 10467348.29 | 11933866.01 | 9509970.982 |
| 4 | 11515669.81 | 10475260.02 | 11671877.22 | 9494500.303 |
| 5 | 12382903.24 | 10931887.31 | 12020628.02 | 9571276.075 |
| Average | 11887628.71 | 10454543.3 | 11723854.51 | 9461237.024 |

Table 6: $J_c$ for A3 after 10 iterations when k = 50.

| Initial | K-means | GA clustering | Improved k-means | IGK |
|---|---|---|---|---|
| 1 | 17552313.95 | 14220466.61 | 16565775.24 | 14149133.94 |
| 2 | 16470491.22 | 15300499.21 | 15220816.69 | 13888135.28 |
| 3 | 15783128.58 | 14620168.73 | 18174366.01 | 14087877.86 |
| 4 | 15541980.51 | 16846546.58 | 15370927.74 | 13543098.84 |
| 5 | 15785254.9 | 14937437.09 | 16165469.41 | 13937084.57 |
| Average | 16226633.83 | 15185023.64 | 16299471.02 | 13921066.1 |

Table 4, 5 and 6 show the calculated error of the K-means algorithm, GA-clustering algorithm, improved K-means algorithm and improved Genetic K-means algorithm, it is obviously that the Squared-error obtained by improved Genetic K-means algorithm for all data set (A1 5376837.128, A2 9461237.024, A3 13921066.1) is better than the calculated

Squared-error by K-means (A1 6776075.178, A2 11887628.71, A3 16226633.83), GA-clustering algorithm (A1 6069568.229, A2 10454543.3, A3 15185023.64) and improved K-means algorithm (A1 15185023.64, A2 11723854.51, A3 16299471.02).

We also run experiments on other dataset is Hepatitis C dataset [2]contains 19 fields with one output field [15]. The output shows whether patients with hepatitis are alive or dead. The purpose of the dataset is to forecast the presence or absence of hepatitis virus given the results of various medical tests carried out on a patient. This database holds 19 attributes. Hepatitis dataset contains 155 samples belonging to two different target classes. The value of K is therefore chosen to be 2 for this data. There are 19 features, 13 binary and 6 attributes with 6–8 discrete values.

Before proceeding for model fitting, we have applied some datareduction techniques in order to trim down the dimensionsbecause the data have the problem of dimensionality's curse.We have applied principal component analysis on the nineteenindependent variables, after which we realized that the first sixprincipal components cover more than

| Initial | K-means | GA clustering | Improved k-means | IGK |
|---|---|---|---|---|
| 1 | 9151.137672 | 9151.137672 | 9398.298872 | 9151.137672 |
| 2 | 9151.137672 | 9810.411534 | 9398.298872 | 9151.137672 |
| 3 | 9151.137672 | 8985.781203 | 9398.298872 | 9151.137672 |
| 4 | 9151.137672 | 9007.065884 | 9398.298872 | 9151.137672 |
| 5 | 9151.137672 | 9056.999657 | 9398.298872 | 9151.137672 |
| Average | 9151.137672 | 9202.27919 | 9398.298872 | 9151.137672 |

98% of the totalvariability of the continuous data space[2].

Therefore, we have acquired six independent variables afterapplying data reduction approaches, these six autonomousvariables are age, bilirubin, Alk phosphate, serum glutamicoxaloacetic transaminase (SGOT), Albumin and prothrombintime (PROTIME).

The results of implementation for Hepatitis C dataset are shown in Table 7.

Table 7: $J_c$ for Hepatitis C after 10 iterations when k = 2.

We have proposed an approach to perform clustering on time series data using GAs and K-means. The proposed approach takes the advantage of the global search capacity of GAs and uses K-means for local fine tuning.

## 8. CONCLUSION

Along with the fast development of database and network, the data scale clustering tasks involved in which becomes more and more large. K-means algorithm is a popular partition algorithm in cluster analysis, which has some limitations when there are some restrictions in computing resources and time, especially for huge size dataset. Genetic algorithm has been used to search for the cluster centers which minimize the clustering error. The improved Genetic K-means algorithm presented in this paper is a solution to handle large scale data, which can select initial clustering center purposefully using GAs, reduce the sensitivity to isolated point, avoid dissevering big cluster, and overcome deflexion of data in some degree that caused by the disproportion in data partitioning owing to adoption of multi-sampling.

## 9. REFERENCES

[1] Yasin, H., JilaniT. A., and Danish, M. 2011. Hepatitis-C Classification using Data Mining Techniques. International Journal of Computer Applications.Vol 24– No.3.

[2] JilaniT. A., Yasin, H., and Yasin,M. M. 2011. PCA-ANN for Classification of Hepatitis-C Patients. International Journal of Computer Applications.Vol14– No.7.

[3] Wang,J. 2006. Encyclopedia of Data Warehousing and Mining. Idea Group Publishing.

[4] Filho,J.L.R.,Treleaven,P.C., andAlippiC. 1994. Genetic Algorithm ProgrammingEnvironments. IEEE Comput, vol.27, pp.28-43.

[5] Maulik, U., and Bandyopadhyay, S. 2000. Genetic Algorithm-BasedClusteringTechnique. Pattern Recognition, vol.33, pp.1455-1465.

[6] Anderberg,M.R. 1973. Cluster Analysis for Application. Academic Press,New York.

[7] Hartigan, J.A.1975. Clustering Algorithms. Wiley, New York.

[8] Devijver,P.A., and Kittler, J. 1982. Pattern Recognition: A Statistical Approach. Prentice-Hall, London.

[9] Jain, A.K., and Dubes, R.C. 1988. Algorithms for Clustering Data. PrenticeHall, Englewood Cliffs, NJ.

[10] Tou, J.T., and Gonzalez, R.C. 1974. Pattern Recognition Principles. Addison Wesley, Reading.

[11] Zhang,Y., Mao,J., and Xiong,Z. 2003: An Efficient Clustering Algorithm.International Conference on Machine Learning and Cybernetics, vol.1,pp.261-265.

[12] Han,J., and Kamber,M. 2000. Data Mining: Concepts and Techniques. Morgan Kaufmann.

[13] Forgy, E.1965. Cluster Analysis of Multivariate Data: Efficiency vs Interpretability of Classifications. Biometrics, pp.21-768.

[14] McQueen,J. 1967. Some methods for classification and analysis of multivariate observations. Computer and Chemistry, vol.4, pp.257-272.

[15] Tang, L., Yang, Z., and Wang, M. 1997. Employing Genetic Algorithm toImprove the K-means Algorithm in

Clustering Analysis.MathematicsStatistics and Application Probability, vol.12.

[16] Cheng,E., Wang,S.,Ning, Y., and Wang, X. 2001. The Design and Realization of Using Representation Point Valid Clustering Algorithm. Patten Recognition and Artificial Intelligence, vol. 14.

[17] Duda, R. O., and Hart, P. E.1973. Patten Classification and Scene Analysis.New York: John Wiley and Sons.

[18] Selim,S. Z., and Alsultan, K. 1991. A Simulated Annealing Algorithm for theClustering Problem. Patten Recognition, vol.24, pp.1003-1008.

[19] Krishna, K., and Narasimha, M. M. 1999. Genetic K-means Algorithm. Systems,IEEE Transactions on Man and Cybernetics, Part B, pp.433-439.

[20] Murthy, C. A., and Chowdhury, N. 1996. In Search of Optimal Clusters usingGenetic Algorithms.PatternRecog. Lett, pp.825-832.

[21] Tang, L., Yang, Z., and Wang, M. 1997. Improve K-means Algorithm of Cluster Method by GA. Mathematical Statistics and Applied Probability,pp.350-356.

[22] Fu, J.,Xu, G., and Wang, Y. 2004. Clustering Based on Genetic Algorithm.Computer engineering, pp.122-124.

[23] Lu, Q., and Yu, J. 2005. K-Means Optimal Clustering Algorithm Based onHybrid Genetic Technique. Journal of East China University of Scienceand Technology (Natural Science Edition), pp.219-222.

[24] Bandyopadhyay, S., and Maulik, U. 2002. An Evolutionary Technique Based onK-Means Algorithm for Optimal Clustering in RN. Information Sciences,pp.221-237.

[25] Yang, S., and Li, Y. 2006.K-means Optimization Study on k Value of K2meansAlgorithm. System Engineering Theory and Application, pp.97-101.

[26] Chittu, V.,Sumathi, N. 2011. A Modified Genetic Algorithm Initializing K-Means Clustering. Global Journal of Computer Science and Technology.

[27] Kumar, V., Chhabra, J.K., and Kumar, D. 2011. Initializing Cluster Center for K-Means Using Biogeography Based Optimization. Advances in Computing, Communication and Control. Vol.125. pp.448-456.

[28] Reddy, D., Mishra,D., and Jana, P. K. 2011. MST-Based Cluster Initialization forK-Means. Advances in Computing. Communication and Control.

[29] Min, W., andSiqing, Y. 2010. Improved K-means clustering based on genetic algorithm. IEEE, Computer Application and System Modeling.

[30] Li, X., Zhang, L., Li, Y., and Wang, Z. 2010.AImproved K-means Clustering Algorithm Combined with the Genetic Algorithm. IEEE, Digital Content,Multimedia Technology and its Applications.

[31] Fayyad, U., Reina,C., and Bradley, P.S. 1998. Initialization of Iterative Refinement Clustering Algorithms. Proceedings of the 4th International Conference on Knowledge Discovery and Data Mining (KDD98), pp.194-198.

[32] Pal, S. K., and Majumder, D. D. 1977. Fuzzy Sets and Decision Making Approaches in Vowel and Speaker Recognition. IEEE Trans. Systems, ManCybernet. SMC, vol.7, pp.625-629.

[33] Fisher, R. A. 1936. The Use of Multiple Measurements in Taxonomic Problems.Ann. Eugenics, vol.3, pp.179-188.

[34] Johnson,R. A., and Wichern, D. W. 1982. Applied Multivariate StatisticalAnalysis. Prentice-Hall, Englewood Cliffs, NJ, 1982.

[35] Virmajoki, O. 2004. Pairwise Nearest Neighbor Method Revisited. PhD thesis,University ofJoensuu, Joensuu, Finland.